\documentclass[runningheads]{llncs}

\input{glyphtounicode}
\usepackage{eccv}

\usepackage{lmodern}

\usepackage{eccvabbrv}

\usepackage{graphicx}
\usepackage{booktabs}
\usepackage{placeins}
\usepackage[accsupp]{axessibility}  

\usepackage{hyperref}

\usepackage{orcidlink}

\begin{document}

\title{RefCompose: Multi-Reference Image Generation via LoRA-Conditioned Diffusion} 

\titlerunning{RefCompose}

\author{Sai Sri Teja Kuppa\inst{1} \and
Parth Shinde\inst{1,2} \and
Priyadharsan Balaji S\inst{1,2} \and
Jinka Harshavardhan\inst{1} \and
Sriprabha Ramanarayanan\inst{1}}

\authorrunning{S.~S.~T.~Kuppa et al.}

\institute{Eros Innovation, India \and
Indian Institute of Technology Madras, India}

\maketitle

\begin{abstract}
  Filmmakers and visual artists routinely need to compose multiple references, actors, locations, props, cultural elements, into a single coherent shot, but existing tools either fail to scale past a handful of references or destroy fine grained subject identity in the process, since per reference tokenization scales memory linearly with reference count $N$ and generated content often departs from the given references rather than reproducing them. We propose \textbf{RefCompose}, a pixel space compositional conditioning framework that decouples \emph{where} things go from \emph{what} they look like, via a single fixed resolution reference canvas that keeps conditioning size constant regardless of reference count. Spatial layout is induced at inference time from a frozen diffusion transformer and extracted via Grounding DINO, requiring no LLM or dedicated layout model, while dual stream LoRA adapters inject a layout derived depth map and the encoded canvas through separate low rank streams, disentangling geometric scaffolding from localized appearance. On the Dense Layout protocol, RefCompose consistently outperforms layout based and state of the art multi reference baselines on color, texture, shape, spatial accuracy, and identity/content preservation at higher reference counts, all with constant inference memory, making it a practical building block for multi subject cinematic composition at production scale.
    \keywords{Diffusion Models \and Image Generation \and LoRA}
  \end{abstract}

\begin{center}
  \includegraphics[width=\linewidth]{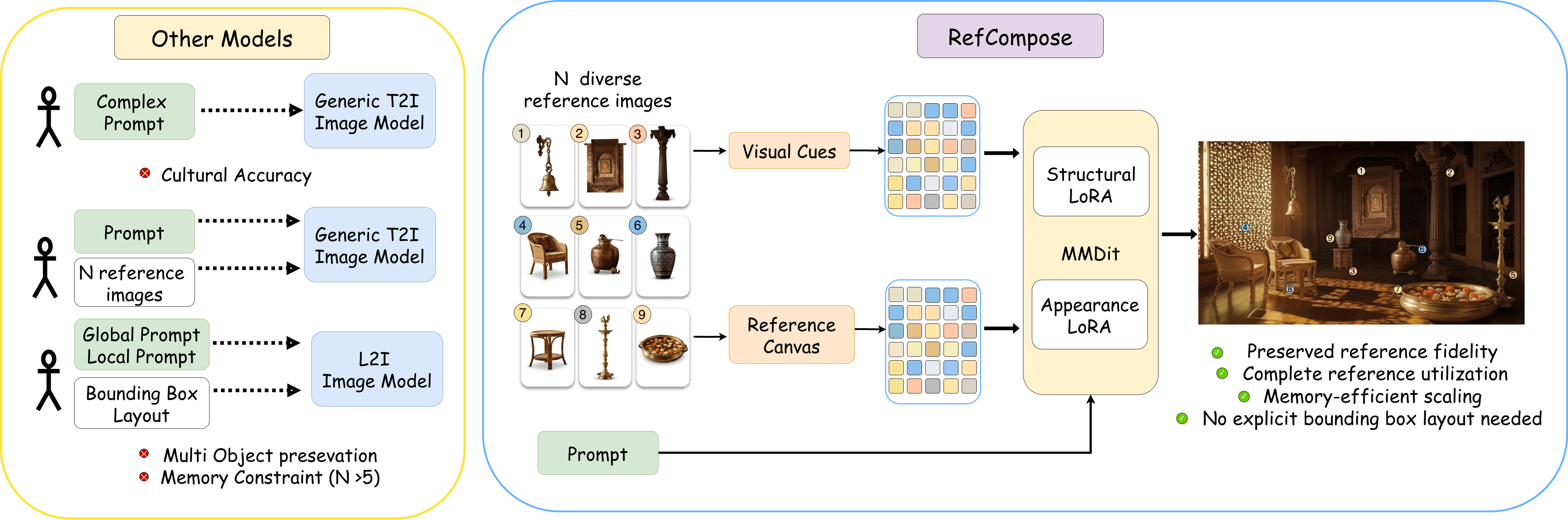}
  \captionof{figure}{Existing approaches suffer from memory overhead scaling linearly with reference count and inability to preserve subject appearance without bounding-box supervision. RefCompose resolves these via a fixed-resolution reference canvas encoding arbitrary references at constant token cost, and layout induced directly from a structured text prompt via dense visual cues.}
  \label{fig:conceptdiagram}
\end{center}

\section{Introduction}
\label{sec:intro}

Controllable image synthesis has emerged as a central challenge in modern generative modelling, with applications spanning advertising, cinematic pre-visualisation, and synthetic data production. 
In the context of movie concept generation using text-to-image (T2I) diffusion models, a single composite frame serves as the critical blueprint across various stages of production, bridging abstract concepts and final visual execution. This visual reference must faithfully reproduce the appearance of specific actors, real-world locations, cultural aesthetics, and narrative objects~\cite{wu2025uno,xverse2024,wang2024msdiffusion,omnigen2_2025}.

Existing T2I methods face significant limitations when textual prompts are the sole control signal. Text prompts often require iterative refinement to encode texture, identity, and style simultaneously across multiple subjects; yet arriving at a prompt that reliably captures all these attributes under heterogeneous data conditions remains a non-trivial challenge. Furthermore, pretrained models underrepresent certain concepts (e.g., cultural), leaving gaps in the common vocabulary due to inherent bias in the training data distribution. This gap undermines the composite image generation through text alone and limits the broader applicability of controllable image generation. Layout-to-Image generative models~\cite{feng2023layoutgpt,li2023gligen,xiang2025instanceassemble,tang2024creatilayout,lian2023layoutdiffusion,zheng2023layoutdiffusion} use bounding-box layouts to pair spatial coordinates with localized prompts alongside a global prompt. However, these sparse region-level descriptions still fail to reconstruct intended appearances under complex occlusion scenarios faithfully. These challenges highlight that dense visual instance cues provide reliable multimodal content, ensuring subject alignment, positioning, and preserving semantic attributes in composite image generation.

Recent reference-based generation methods scale to multiple subjects, addressing cultural and ethnic fidelity gaps left by text-only approaches~\cite{wu2025uno,he2025dreamo,xiao2024omnigen,omnigen2_2025,wang2024instantid,li2024photomaker}, but encoding references as independently concatenated token sequences causes token budget to grow linearly, becoming prohibitive beyond four to six images, especially at high resolution~\cite{ye2023ip,wu2025uno,xverse2024,xiao2024omnigen,omnigen2_2025}. Attention-based alternatives like QKV injection avoid this but remain spatially imprecise, with appearance leaking across object boundaries and inversion-based reconstruction deviating from the reference even at conservative noise levels~\cite{hertz2022prompt,mokady2023null,tumanyan2023plug}. We observe that token-scaling stems from the representation, not multi-reference generation itself: RefCompose instead compresses any number of reference images onto a single fixed-resolution canvas in pixel space, keeping token count constant (Figure~\ref{fig:conceptdiagram}). A coarse prompt-induced layout from the frozen diffusion transformer yields a monocular depth map for geometric guidance without extra supervision; the canvas and depth map, encoding appearance and structure respectively, are injected via a LoRA-based conditioning framework for parameter-efficient adaptation. Our contributions:

1. \textbf{RefCompose}, a multimodal framework combining spatial compositional conditioning and textual prompts for multi-reference generation, via a \textbf{canvas-based multi-reference representation} that eliminates token explosion at constant memory cost.

2. A \textbf{decoupled structure-appearance conditioning} scheme where a depth map controls global geometry and the reference canvas injects localized appearance, both aligned to the same spatial layout.

3. A \textbf{prompt-induced layout}, where a generic composition prompt induces a coarse multi-subject layout from a frozen diffusion transformer with no bounding-box supervision or dedicated layout model, and a depth map derived from it provides geometric guidance throughout denoising.

\section{Related Work}
\label{sec:related}

\paragraph{Reference-guided image synthesis.}
Conditioning generation on reference images has a long history in diffusion literature~\cite{zhang2023adding,xverse2024,xiang2025instanceassemble,he2025dreamo,tang2024creatilayout,kumari2023multiconcept,ruiz2023dreambooth}.
IP-Adapter~\cite{ye2023ip} couples image and text conditioning via cross-attention, enabling subject-driven generation from a single reference, and follow-up methods share a common structure of encoding each reference independently and injecting it as an additional token sequence~\cite{wang2024instantid, li2024photomaker, ruiz2023dreambooth, kumari2023multiconcept}, so token cost grows with reference count, an issue made explicit by architectures that concatenate multiple VAE encodings, which overflow GPU memory beyond a small number of references~\cite{ye2023ip, wang2024msdiffusion, wu2025uno, xverse2024, xiao2024omnigen, omnigen2_2025}.
Our canvas-based representation avoids this by keeping token count constant regardless of reference count.

\paragraph{Layout-conditioned generation.}
Spatial control has been pursued via layout priors of various kinds~\cite{zhang2023adding, li2023gligen, feng2023layoutgpt, lian2023layoutdiffusion, zheng2023layoutdiffusion, tang2024creatilayout, xiang2025instanceassemble}: LayoutGPT~\cite{feng2023layoutgpt} uses an LLM to generate bounding boxes for conditioning, LayoutDiffusion~\cite{zheng2023layoutdiffusion} models spatial arrangement distributions directly, and GLIGEN~\cite{li2023gligen} injects grounding tokens at specified locations via gated self-attention for object-level placement.
These add pipeline complexity; we find an explicit layout component unnecessary, since state-of-the-art DiTs can use a generic composition prompt for coarse layout sufficient for canvas placement and depth extraction; a comparison against specialized layout generators on COCO-GR is provided in the supplementary material.

\paragraph{Depth, structural, and LoRA-based conditioning.}
ControlNet~\cite{zhang2023adding} showed pre-computed spatial signals (depth, canny edges, pose) provide strong structural guidance for UNet-based diffusion via a trainable parallel encoder, though adapting this to DiTs is non-trivial given the architectural shift from UNet.
EasyControl~\cite{zhang2025easycontrol} bridges this gap with a lightweight Condition Injection LoRA integrating spatial conditioning into DiT attention via causal attention and KV caching, achieving plug-and-play control without retraining; we build on EasyControl as the conditioning backbone, using depth maps from layout-induced generations for geometric guidance while leaving base weights intact.
More broadly, LoRA~\cite{hu2022lora} is the standard mechanism for efficient fine-tuning, injecting style, identity, and appearance features without modifying base parameters, and offers more stable transfer than attention manipulation strategies like QKV injection or self-attention swapping, which are sensitive to target object footprint and inconsistent for small subjects~\cite{hertz2022prompt, mokady2023null, tumanyan2023plug}.
We also find inversion-based pipelines fragile: above 0.5 noise strength, structural distortion and fidelity loss become severe.
EasyControl's Condition Injection LoRA avoids both failure modes via rank-decomposed weight updates rather than activation manipulation.

\paragraph{Multi-condition generation.}
Handling multiple conditioning signals is non-trivial since signals may conflict in attention layers.
Most closely related is Canvas-to-Image~\cite{dalva2025canvastoimage}, which performs compositional generation from a multimodal canvas but lacks explicit orientation control and degrades in densely cluttered scenes; Mixture-of-Experts adapters~\cite{wu2024mole} and composable diffusion~\cite{liu2022compositional} address multi-signal conditioning via routing or score composition, typically at the cost of inference complexity or per-condition fidelity.
We instead adopt a disentangled design: EasyControl~\cite{zhang2025easycontrol} separates spatial and subject signals via position-aware training and KV caching, without mutual interference, and our method exploits this to jointly inject depth (structure) and canvas (appearance) within a single forward pass.

\section{Methodology}
\label{sec:method}

\begin{figure}[!t]
  \centering
  \includegraphics[width=\linewidth]{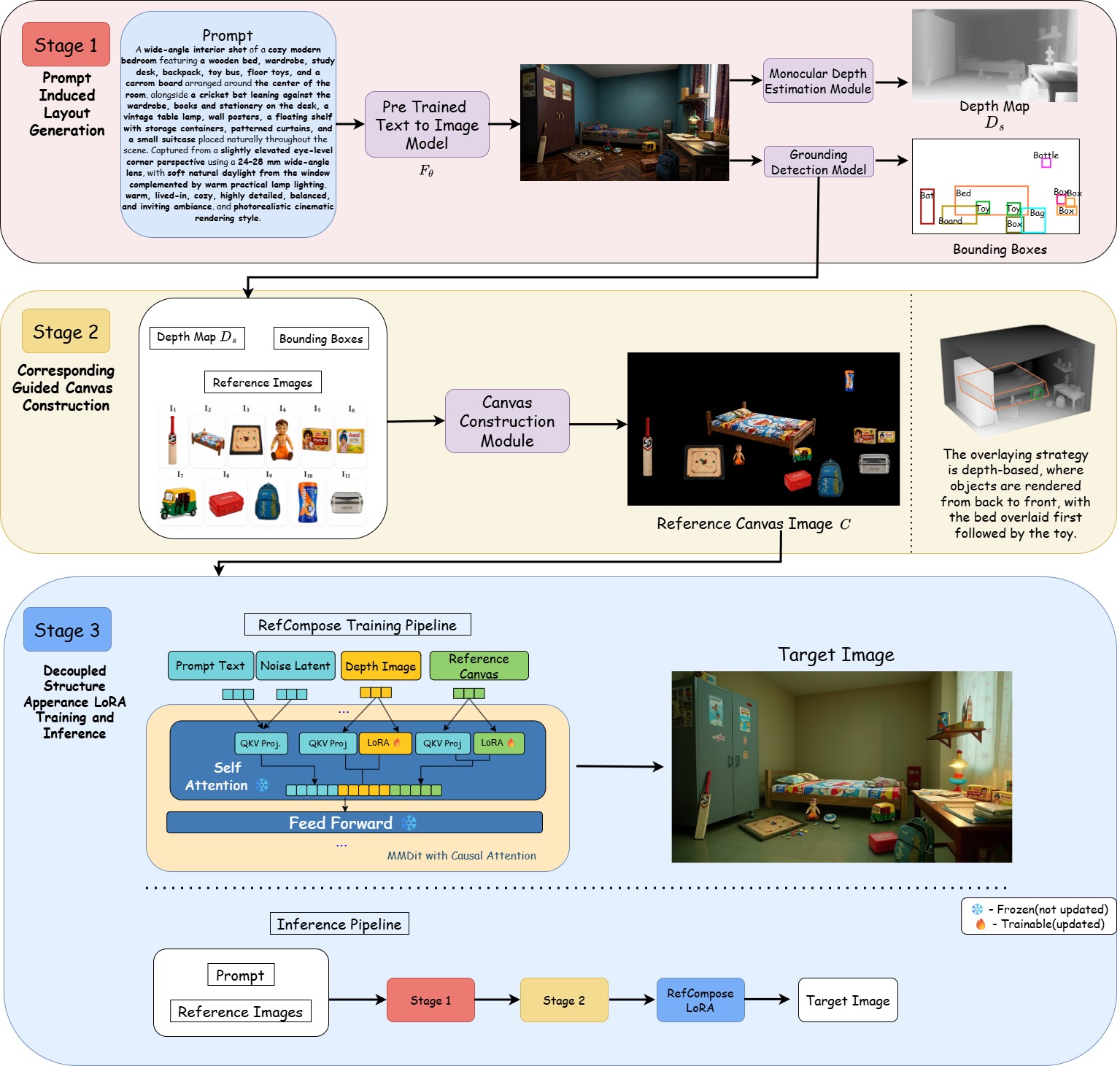}
  \caption{Overview of RefCompose: prompt-induced layout generation, correspondence-guided canvas construction, and decoupled structure-appearance conditioning via dual LoRA streams in the MMDiT blocks, preserving full-resolution identity at constant token complexity.}
  \label{fig:pipeline}
\end{figure}

Multi-reference image generation must preserve each reference's identity while keeping conditioning cost fixed regardless of reference count.
Token-concatenation approaches fail here, since each added reference appends a new encoding sequence, causing linear growth in attention complexity and cross-reference interference.
We decompose controllable generation into two orthogonal sub-problems: \emph{where} each object appears (structure) and \emph{what} it looks like (appearance).
Structure is encoded via a layout-derived geometric scaffold; appearance is grounded locally via a fixed-resolution reference canvas.
Both signals are injected into a frozen diffusion transformer via decoupled LoRA streams, keeping token complexity constant regardless of reference cardinality.

\paragraph{Notation.}
Let $\mathcal{R} = \{I_1, \ldots, I_N\}$ be $N$ RGB references, $\mathbf{p}$ an object-centric prompt with positional language, and $(H,W)$ the output resolution.
$\mathcal{F}_\theta$ is the frozen diffusion transformer under $T{=}40$ flow-matching steps; $\mathcal{E}, \mathcal{E}^{-1}$ the VAE encoder/decoder; and $r$ the LoRA rank (distinct from the reference set $\mathcal{R}$).

\subsection{Prompt-Induced Layout Generation}
\label{sec:layout}
Rather than invoking an external layout model or LLM, we derive spatial structure from the frozen diffusion transformer $\mathcal{F}_\theta$ itself.
We restrict $\mathbf{p}$ to compositional and relational language (scene type, camera framing, and inter-subject placement; \emph{e.g.}, ``a person on the left, a product on the right'') and delegate photometric identity to the reference canvas (Section~\ref{sec:canvas}).
This splits conditioning into orthogonal channels: $\mathbf{p}$ specifies \emph{where} subjects belong; the canvas specifies \emph{what} they look like.
Text-conditioned flow-matching over $T{=}40$ denoising steps yields a coarse layout image:
\begin{equation}
  L = \mathcal{F}_\theta(\mathbf{p}) \in \mathbb{R}^{H \times W \times 3}.
\end{equation}
Open-vocabulary detector $\mathcal{G}$ (Grounding DINO~\cite{liu2023grounding}) localises each reference in $L$, producing boxes that tie references to output regions:
\begin{equation}
  \mathcal{B} = \mathcal{G}(L) = \{b_1, \ldots, b_N\}, \quad b_k = (x_k, y_k, w_k, h_k),
\end{equation}
where $b_k$ is the region assigned to reference $I_k$.
Because depth $D_s=\Psi(L)$ (Section~\ref{sec:depth}) and canvas assembly (Section~\ref{sec:canvas}) both read from the same $L$ and $\mathcal{B}$, geometric and appearance conditioning share one spatial frame, avoiding misalignment that would arise if layout, boxes, and depth were computed independently.

\subsection{Reference Canvas Representation}
\label{sec:canvas}
Sequential encoding (independently tokenising and concatenating each reference) scales linearly with $N$ and forces incompatible token streams at every attention layer.
We instead use a \emph{reference canvas} $\mathcal{C} \in \mathbb{R}^{H \times W \times 3}$: a fixed-resolution image in which each reference occupies its expected output region, built in pixel space to preserve texture and identity.
Each reference is resized and placed at its box:
\begin{equation}
  \phi(I_k, b_k) = \mathrm{Resize}(I_k,\; h_k \times w_k)\;\text{placed at}\; b_k.
\end{equation}
When boxes overlap, the nearer reference (by depth, Section~\ref{sec:depth}) overwrites the farther one.
$\mathcal{C}$ is encoded once by the VAE:
\begin{equation}
  z_{\mathcal{C}} = \mathcal{E}(\mathcal{C}) \in \mathbb{R}^{(H/f) \times (W/f) \times c},
\end{equation}
with $f$ the VAE downsampling factor ($f=8$ typically) and $c$ the latent channel dimension.
Thus $|z_{\mathcal{C}}| = \text{const}$ for all $N$, versus sequential schemes costing $N \cdot (H/f)(W/f)c$, capping GPU memory independent of reference count.

\subsection{Geometric Scaffolding via Depth Estimation}
\label{sec:depth}
Compositing alone lacks depth ordering, surface orientation, and occlusion cues needed for 3D plausibility. We estimate a monocular depth map from the layout image:
\begin{equation}
  D_s = \Psi(L) \in \mathbb{R}^{H \times W},
\end{equation}
using Depth Anything 3~\cite{depthanything3}. Deriving $D_s$ from $L$ (rather than the references) ensures $L$, $\mathcal{B}$, and $D_s$ share an exact spatial frame. Since $\Psi$ runs on generated layouts, not real photos, we treat its output as a relative geometric prior, not metric ground truth.

For each reference $k$, we compute mean depth $\bar{d}_k = \mathrm{mean}(D_s \mid b_k)$ and sort far-to-near as permutation $\pi$. The canvas is assembled by sequential overwrite compositing:
\begin{equation}
  \mathcal{C} =
  \mathrm{Paste}\!\left(
    \phi(I_{\pi(1)}, b_{\pi(1)}), \ldots,
    \phi(I_{\pi(N)}, b_{\pi(N)})
  \right),
\end{equation}
overwriting overlaps with the nearer reference, a practical ordering signal, not an infallible occlusion oracle.

$D_s$ is VAE-encoded for compatibility with latent denoising:
\begin{equation}
  z_{D_s} = \mathcal{E}(D_s) \in \mathbb{R}^{(H/f) \times (W/f) \times c},
\end{equation}
carrying occlusion order, surface orientation, and spatial extent in a form commensurate with $z_{\mathcal{C}}$.

\subsection{Decoupled Structure-Appearance (DSA) LoRA}
Naively concatenating $z_{D_s}$ and $z_{\mathcal{C}}$ lets depth features leak into appearance and vice versa. We instead route each through a dedicated low-rank module.

For frozen projection $W \in \mathbb{R}^{d_{\mathrm{out}} \times d_{\mathrm{in}}}$ in the double-stream attention blocks, each LoRA branch learns:
\begin{equation}
  \Delta W_m = B_m A_m, \qquad
  A_m \in \mathbb{R}^{r \times d_{\mathrm{in}}}, \quad
  B_m \in \mathbb{R}^{d_{\mathrm{out}} \times r}, \quad
  m \in \{s,a\},
\end{equation}
with rank $r$, $s$ the structure branch ($z_{D_s}$), and $a$ the appearance branch ($z_{\mathcal{C}}$). Effective projection: $W + \Delta W_s + \Delta W_a$, base $W$ frozen.

Starting from terminal latent $z_T$, flow-matching sampling yields
\begin{equation}
  \hat{z}_0 = \mathcal{F}_\theta\bigl(\mathbf{p}, z_T, z_{\mathcal{C}},
  z_{D_s}; \Delta W_s + \Delta W_a\bigr), \quad T{=}40,
\end{equation}
where $\mathbf{p}$ provides semantic grounding, $z_{D_s}$ the
geometric scaffold, and $z_{\mathcal{C}}$ localised appearance, and
\begin{equation}
  v_\theta\bigl(z_t, t, \mathbf{p}, z_{\mathcal{C}}, z_{D_s};
  \Delta W_s + \Delta W_a\bigr)
\end{equation}
denotes the learned velocity at step~$t$. Output via VAE decoding:
\begin{equation}
  \hat{I} = \mathcal{E}^{-1}(\hat{z}_0) \in \mathbb{R}^{H \times W \times 3}.
\end{equation}

For training, let $I$ denote the ground-truth target image and $z_0 = \mathcal{E}(I)$ its latent encoding. We sample noise $z_1 \sim \mathcal{N}(0, I)$, timestep $t \sim \mathcal{U}[0, 1]$, and form the interpolated latent $z_t = (1 - t) z_0 + t z_1$. The LoRA adapters are trained by regressing $v_\theta$ onto the constant rectified-flow velocity $z_1 - z_0$, i.e.\ by minimising the conditional flow-matching objective
\begin{equation}
  \mathcal{L}_{\text{FM}} = \mathbb{E}_{t,\, z_0,\, z_1} \left[ \left\lVert v_\theta\bigl(z_t, t, \mathbf{p}, z_{\mathcal{C}}, z_{D_s}; \Delta W_s + \Delta W_a\bigr) - (z_1 - z_0) \right\rVert_2^2 \right],
\end{equation}
where only $\Delta W_s + \Delta W_a$ is updated while $\mathcal{F}_\theta$ remains frozen. At inference, sampling starts from $z_T \sim \mathcal{N}(0, I)$ and integrates $v_\theta$ over $T{=}40$ steps to obtain $\hat{z}_0$ and decoded output $\hat{I}$.

\section{Experiments}

\subsection{Dataset Preparation}
\label{sec:dataset-prep}

\textbf{Dataset Construction.} We train on ultra high resolution cinematic imagery at $1280 \times 720$ resolution, drawn from a private dataset sourced from movies, where most object references are canonicalized to a neutral pose. We complement this with a synthetic dataset generated using FLUX.2~\cite{blackforestlabs2025flux2}, providing diverse object positions and orientations so the model learns pose variation while the real dataset supplies richer textures; further details are provided in the supplementary material. Our pipeline proceeds in five stages. First, each movie shot is passed through Qwen3-VL~\cite{qwen3vl} to enumerate all objects present in the scene (actors, props, and landmarks). Second, Grounding DINO~\cite{liu2023grounding} takes these object descriptions as text prompts to detect corresponding bounding boxes, retaining only shots yielding more than four detections to ensure compositional density. Third, each detected crop is fed to FLUX.2 Dev~\cite{blackforestlabs2026flux2dev} to synthesize a neutral-pose, neutral-lighting canonicalization, removing scene-specific pose and illumination variation. Fourth, we filter canonicalized outputs against their original crops using DINO~\cite{caron2021dino} cosine similarity, retaining a sample only if similarity exceeds $0.90$. Finally, retained shots are annotated again with Qwen3-VL, producing both scene-level and crop-level descriptions (Figure~\ref{fig:data_pipeline}). This pipeline yields approximately $50{,}000$ samples drawn from $1{,}000$ distinct films, together with additional synthetic samples. The final training corpus is evenly split between private cinematic and synthetic samples (50\%/50\%).
\begin{center}
  \includegraphics[width=\linewidth]{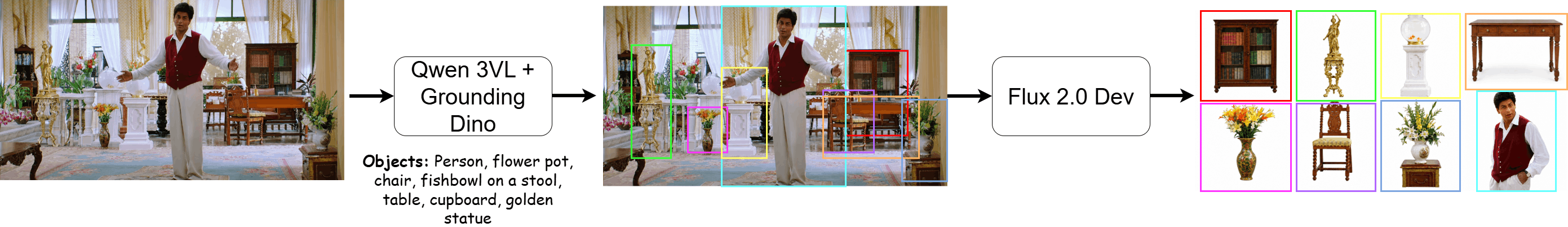}
  \captionof{figure}{Overview of the dataset preparation pipeline. Objects are identified using Qwen3-VL, localized with Grounding DINO, canonicalized to a neutral pose and lighting using FLUX.2 Dev~\cite{blackforestlabs2026flux2dev}, filtered via DINO~\cite{caron2021dino} similarity, and finally annotated by Qwen3-VL.}
  \label{fig:data_pipeline}
\end{center}

\subsection{Inference Prompt Template}
\label{sec:layout-gen}

We instantiate the layout pipeline of Section~\ref{sec:layout} at $1280 \times 720$ with FLUX.2 Dev~\cite{blackforestlabs2026flux2dev} as $\mathcal{F}_\theta$.
For fair comparison, RefCompose and all baselines share one fixed template; bracketed slots are populated automatically from Qwen3-VL~\cite{qwen3vl} scene and crop descriptions during benchmark evaluation (Section~\ref{sec:quant}):
\begin{quote}
\small
\textit{A [shot type] of a [scene/environment] featuring [primary objects] arranged around [central subject], alongside [secondary objects/details] placed naturally throughout the scene. Captured from a [camera perspective] using a [lens type], with [lighting condition], [mood/style descriptors], and [photorealistic/cinematic rendering style].}
\end{quote}

Because box sizes in $L$ are prompt-driven, relative subject prominence can be steered through shot type and scale descriptors in the template.
When $N$ is large, each $b_k$ covers a smaller canvas fraction, so $\phi(I_k, b_k)$ downsamples references into tighter patches and within-canvas identity detail degrades even though $|z_{\mathcal{C}}|$ stays fixed; we assign larger relational scale to priority subjects in $\mathbf{p}$ so they receive proportionally larger boxes in $L$.

\subsection{RefCompose LoRA Training}
We extend EasyControl~\cite{zhang2025easycontrol}'s Condition Injection LoRA to inject depth (structure) and a reference-canvas latent $z_{\mathcal{C}}$ (appearance) into a frozen FLUX.2 Dev diffusion transformer~\cite{blackforestlabs2026flux2dev} via separate low-rank updates $\Delta W = BA$ with rank $r{=}32$ in the double-stream attention blocks; causal attention and KV caching keep the two streams disentangled, with the subject LoRA loaded before the depth LoRA at inference.

Training runs for $50{,}000$ iterations at $1280 \times 720$ in three stages while base weights remain frozen: (1) $10{,}000$ iterations on depth with a black canvas, initialising the structure branch before appearance cues are introduced; without this warm-up the model tends to use the canvas stream to compensate for geometry, reducing depth/appearance separation; (2) $20{,}000$ iterations with crops placed at Grounding DINO~\cite{liu2023grounding} boxes on the ground-truth image, teaching appearance grounding under clean placements; (3) the remaining iterations add photometric jitter and mild spatial perturbations (affine warps, small translations) to crop placement, improving robustness to noisier boxes from the inference-time layout pipeline. Because both branches condition on a single encoded canvas rather than per-reference token streams, GPU memory at inference stays approximately constant with reference count.

\subsection{Quantitative Analysis}
\label{sec:quant}

\paragraph{Benchmark.}
We evaluate on the \emph{Dense Layout} test split of InstanceAssemble~\cite{xiang2025instanceassemble}: $5{,}000$ held-out composites ($\sim$8.3 annotated objects/image across people, props, environments, and product-like assets) with ground-truth boxes and per-instance text descriptions.
All metrics in Tables~\ref{tab:memory} and~\ref{tab:quant_results} are computed on this split unless stated otherwise.

\paragraph{Evaluation protocol and GPU memory.}
All methods run on a single NVIDIA H100 at \(1280 \times 720\) using the prompt template from Section~\ref{sec:layout-gen}. For each image, we select five object crops from the largest ground-truth boxes by area and canonicalize them via FLUX.2 Dev to a neutral pose, viewpoint, and lighting, consistent with the training data construction procedure (Section~\ref{sec:dataset-prep}); identity must therefore be reconstructed from this re-rendered appearance rather than copied from the original target pixels. Scene prompts are generated with Qwen3-VL~\cite{qwen3vl}, so that all methods receive identical textual and visual information. CreatiLayout and InstanceAssemble~\cite{xiang2025instanceassemble} receive only Dense Layout boxes and text, while UNO~\cite{wu2025uno}, XVerse~\cite{xverse2024}, and RefCompose additionally receive the canonicalized reference crops. Among all baselines, UNO and XVerse are the most architecturally compatible with RefCompose and achieve the strongest reference-image performance, so we benchmark primarily against them. Region-wise metrics use object localizations from Grounding DINO~\cite{liu2023grounding}.

As Table~\ref{tab:memory} shows, token-based methods scale in memory with reference count and go out-of-memory at \(N=6\), whereas RefCompose remains approximately constant via canvas compression. Baselines are therefore reported at \(N{=}5\), while Table~\ref{tab:quant_results} additionally reports RefCompose at \(N{>}5\) to demonstrate scalability without added memory cost. In this \(N{>}5\) setting, the reference count follows the per-image annotation count from Dense Layout (capped at \(N=15\)); Table~\ref{tab:quant_results} thus averages over image-varying \(N\) for RefCompose, whereas baseline rows use a uniform cap of \(N{=}5\).

Following LayoutSAM-Eval~\cite{xiang2025instanceassemble}, we report \emph{region-wise} quality (Spatial, Color, Texture, Shape: placement accuracy and appearance/structure fidelity) and \emph{global-wise} quality (PICK~\cite{kirstain2023pickapic}, CLIP-T~\cite{radford2021clip} for image-text alignment; DINO~\cite{caron2021dino} for composition similarity; DINO$_{\text{refs}}$ for reference consistency; CLIP-I~\cite{radford2021clip} for image-level similarity). Higher is better throughout.

\begin{table}[t]
  \centering
  \caption{Peak GPU memory (GB) at inference for varying reference count $N$ (batch size $1$; single H100; $1280 \times 720$). UNO~\cite{wu2025uno} and XVerse~\cite{xverse2024} scale with reference count and fail at $N{=}6$; RefCompose stays constant via canvas compression. OOM = out-of-memory.}
  \label{tab:memory}
  \setlength{\tabcolsep}{5pt}
  \begin{tabular}{lccc}
    \toprule
    \textbf{Method} & $N{=}2$ & $N{=}4$ & $N{=}6$ \\
    \midrule
    CreatiLayout~\cite{tang2024creatilayout} & 34.7 & 34.8 & 34.9 \\
    InstanceAssemble~\cite{xiang2025instanceassemble} & 35.2 & 35.3 & 35.4 \\
    UNO~\cite{wu2025uno} & 60.1 & 74.8 & OOM \\
    XVerse~\cite{xverse2024} & 59.6 & 75.3 & OOM \\
    \midrule
    \textbf{RefCompose} & \textbf{69.0} & \textbf{69.0} & \textbf{69.0} \\
    \bottomrule
  \end{tabular}
\end{table}

\begin{table*}[!t]
  \centering
  \caption{Quantitative comparison on the Dense Layout test set~\cite{xiang2025instanceassemble} ($5{,}000$ images; mean $8.3$ objects/image). Baselines use five references (UNO/XVerse cannot run beyond this on one H100); the $N{>}5$ RefCompose row uses additional references. Higher is better. \textbf{Bold} = best overall; improvement row compares RefCompose ($N{>}5$) against the strongest reference-image baseline per metric.}
  \label{tab:quant_results}
  \resizebox{\textwidth}{!}{%
  \setlength{\tabcolsep}{8pt}
  \begin{tabular}{lccccccccc}
    \toprule
    {\textbf{Method}}
    & \multicolumn{4}{c}{\textbf{Region-wise Quality}}
    & \multicolumn{5}{c}{\textbf{Global-wise Quality}} \\
    \cmidrule(lr){2-5} \cmidrule(lr){6-10}
    & \textbf{Spatial}$\uparrow$
    & \textbf{Color}$\uparrow$
    & \textbf{Texture}$\uparrow$
    & \textbf{Shape}$\uparrow$
    & \textbf{PICK}$\uparrow$
    & \textbf{DINO}$\uparrow$
    & \textbf{DINO$_{\text{refs}}$}$\uparrow$
    & \textbf{CLIP-I}$\uparrow$
    & \textbf{CLIP-T}$\uparrow$ \\
    \midrule
    \multicolumn{10}{l}{\textit{Layout-only baselines}} \\
    CreatiLayout~\cite{tang2024creatilayout}
      & 0.726 & 0.487 & 0.518 & 0.501
      & 0.222 & 0.696 & 0.482 & 0.727 & \textbf{0.335} \\
    InstanceAssemble~\cite{xiang2025instanceassemble}
      & 0.790 & 0.590 & 0.620 & 0.610
      & 0.221 & 0.686 & 0.457 & 0.703 & 0.330 \\
    \midrule
    \multicolumn{10}{l}{\textit{Reference-image baselines}} \\
    UNO~\cite{wu2025uno}
      & 0.750 & 0.500 & 0.530 & 0.520
      & 0.225 & 0.821 & 0.548 & 0.742 & 0.329 \\
    XVerse~\cite{xverse2024}
      & 0.665 & 0.421 & 0.442 & 0.428
      & 0.218 & 0.686 & 0.506 & 0.731 & 0.302 \\
    \midrule
    \textbf{RefCompose} ($N{=}5$)
      & 0.908 & 0.826 & 0.854 & 0.849
      & 0.229 & 0.959 & 0.613 & 0.766 & 0.327 \\
    \textit{vs. best reference baseline}
      & \textcolor{green!60!black}{+0.172}
      & \textcolor{green!60!black}{+0.338}
      & \textcolor{green!60!black}{+0.336}
      & \textcolor{green!60!black}{+0.341}
      & \textcolor{green!60!black}{+0.006}
      & \textcolor{green!60!black}{+0.150}
      & \textcolor{green!60!black}{+0.077}
      & \textcolor{green!60!black}{+0.036}
      & 0.000 \\
    \textbf{RefCompose} ($N{>}5$)
      & \textbf{0.922} & \textbf{0.838} & \textbf{0.866} & \textbf{0.861}
      & \textbf{0.231} & \textbf{0.971} & \textbf{0.625} & \textbf{0.778} & 0.329 \\
    \bottomrule
  \end{tabular}%
  }
\end{table*}

\begin{figure}[t]
\centering
\includegraphics[width=\linewidth]{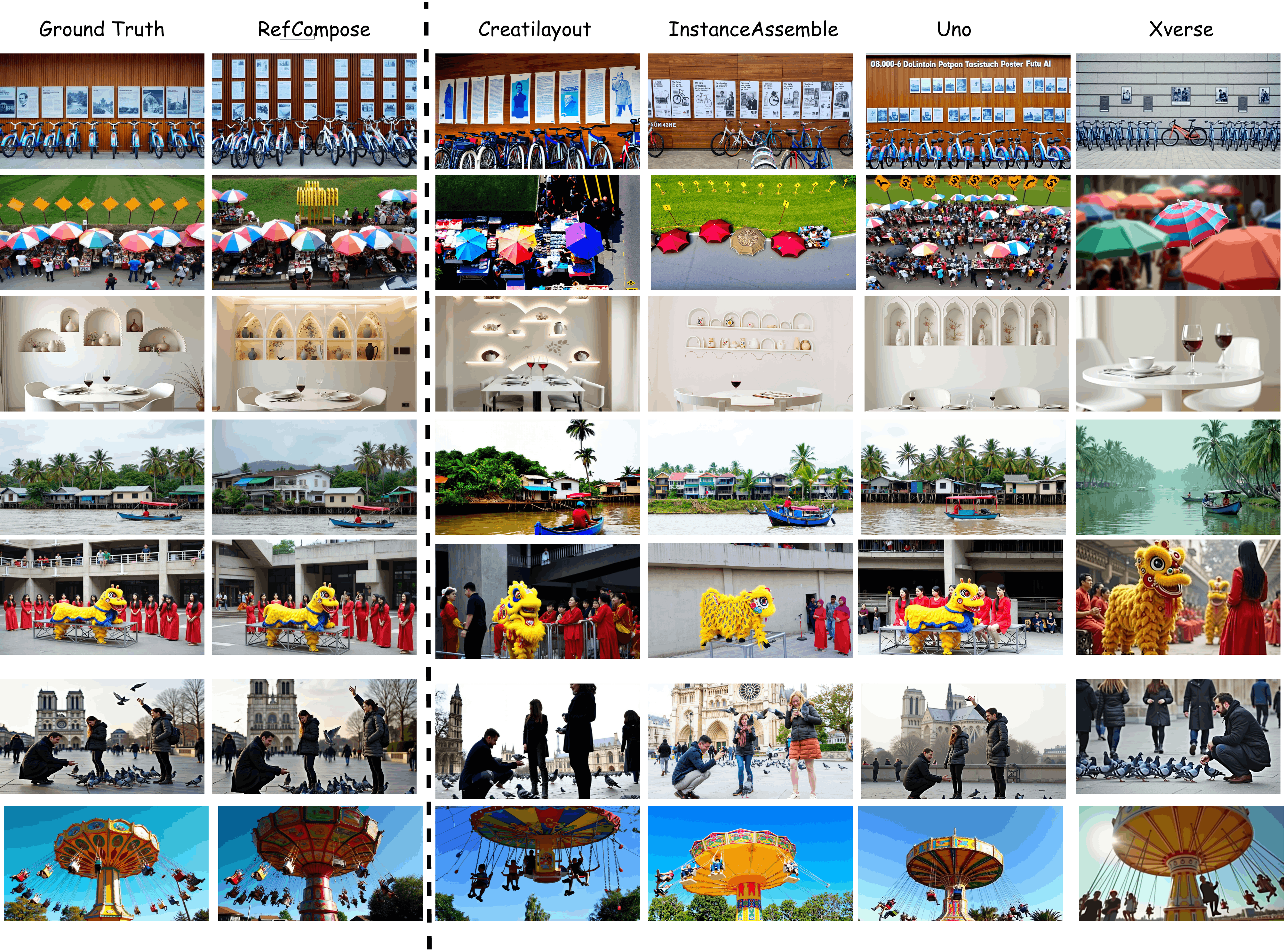}
\caption{Qualitative comparison of RefCompose against CreatiLayout, InstanceAssemble, UNO, and XVerse across seven multi-subject scenes (ground truth leftmost). Layout-conditioned baselines hallucinate appearance; token-based methods show subject drift and detail loss under multi-object composition. RefCompose reproduces both spatial arrangement and photometric identity most faithfully.}
\label{fig:teaser}
\end{figure}

Layout-conditioned baselines (CreatiLayout, InstanceAssemble) receive no visual reference, so photometric metrics mainly reflect the value of reference conditioning; UNO~\cite{wu2025uno} and XVerse~\cite{xverse2024} are the more architecturally matched comparisons.

\paragraph{Comparison to token-based reference methods.}
RefCompose outperforms token-based methods UNO~\cite{wu2025uno} and XVerse~\cite{xverse2024}, with higher spatial accuracy (0.908 vs.\ 0.750), colour fidelity (0.826 vs.\ 0.500), and reference alignment (DINO$_\text{refs}$: 0.613 vs.\ 0.548). Canvas conditioning also holds memory roughly constant, whereas token concatenation scales linearly with reference count, underscoring the benefit of decoupling appearance and layout cues in multi-object generation.

\paragraph{Comparison to layout-conditioned methods.}
CreatiLayout and InstanceAssemble score well on text alignment, with CreatiLayout achieving the top CLIP-T ($0.335$) and InstanceAssemble the higher spatial score ($0.790$), showing that explicit box supervision aids placement even without reference pixels. Their much lower colour, texture, and shape scores reflect the absence of reference pixels rather than an architectural limitation, since appearance must be derived from text alone. RefCompose's photometric and DINO$_\text{refs}$ gains are thus best judged against UNO and XVerse, which share the same memory-bounded regime and reference crops.

\paragraph{Global preference and text alignment.}
PICK scores are tight (0.218 for XVerse to 0.229 for RefCompose), making the larger spatial, photometric, and DINO$_\text{refs}$ gaps more informative. RefCompose's CLIP-T ($0.327$) is slightly below UNO ($0.329$) and the layout-only baselines, suggesting text-only alignment is complementary to, not the primary indicator of, reference-grounded composition quality.

\subsection{Qualitative Analysis}
Figure~\ref{fig:teaser} corroborates the quantitative trends across seven multi-subject scenes.
CreatiLayout and InstanceAssemble place objects coherently from boxes and text but hallucinate colour, texture, and identity, especially for distinctive materials where language is a poor substitute for pixels.
UNO~\cite{wu2025uno} and XVerse~\cite{xverse2024} transfer some reference appearance yet show subject drift and softened fine detail under multi-object layouts in our qualitative examples.
RefCompose composites references at full resolution on a shared canvas and conditions denoising with an aligned depth map, yielding subjects at correct locations with faithful photometric detail; overlapping instances are best separated where depth discontinuities match canvas boundaries and the dual LoRA streams limit cross-region leakage.
We do not include Canvas-to-Image~\cite{dalva2025canvastoimage} in this comparison because its source code is not publicly available and it lacks disentangled dual-LoRA control over structure and appearance.
Our ablation confirms that neither modality alone suffices, and that merging depth and canvas through a single LoRA stream causes cross-signal overwriting that degrades either geometric placement or textural fidelity.
The lower CLIP-T score relative to layout-conditioned baselines reflects an expected tradeoff: canvas conditioning can shift generation toward reference fidelity at a minor cost to global text alignment, so spatial and DINO$_\text{refs}$ metrics are more diagnostic for reference-grounded composition.
Residual failures appear when prompt-induced layouts imply ambiguous occlusion or implausible depth, which bounds geometric plausibility even though the depth prior corrects most floating and interpenetration artefacts relative to layout-only baselines.

\paragraph{Multi-reference flexibility.}
Figure~\ref{fig:multi_reference} evaluates two distinct reference sets for the same scene configuration.
A FLUX.2~\cite{blackforestlabs2025flux2}-generated layout image defines where each subject should appear, while the estimated depth map supplies additional geometric cues for relative orientation and placement.
By conditioning jointly on the reference canvas and depth prior, RefCompose synthesises coherent outputs for both reference sets, preserving the intended spatial arrangement while transferring the appearance of each supplied subject.
\begin{center}
  \includegraphics[width=\linewidth]{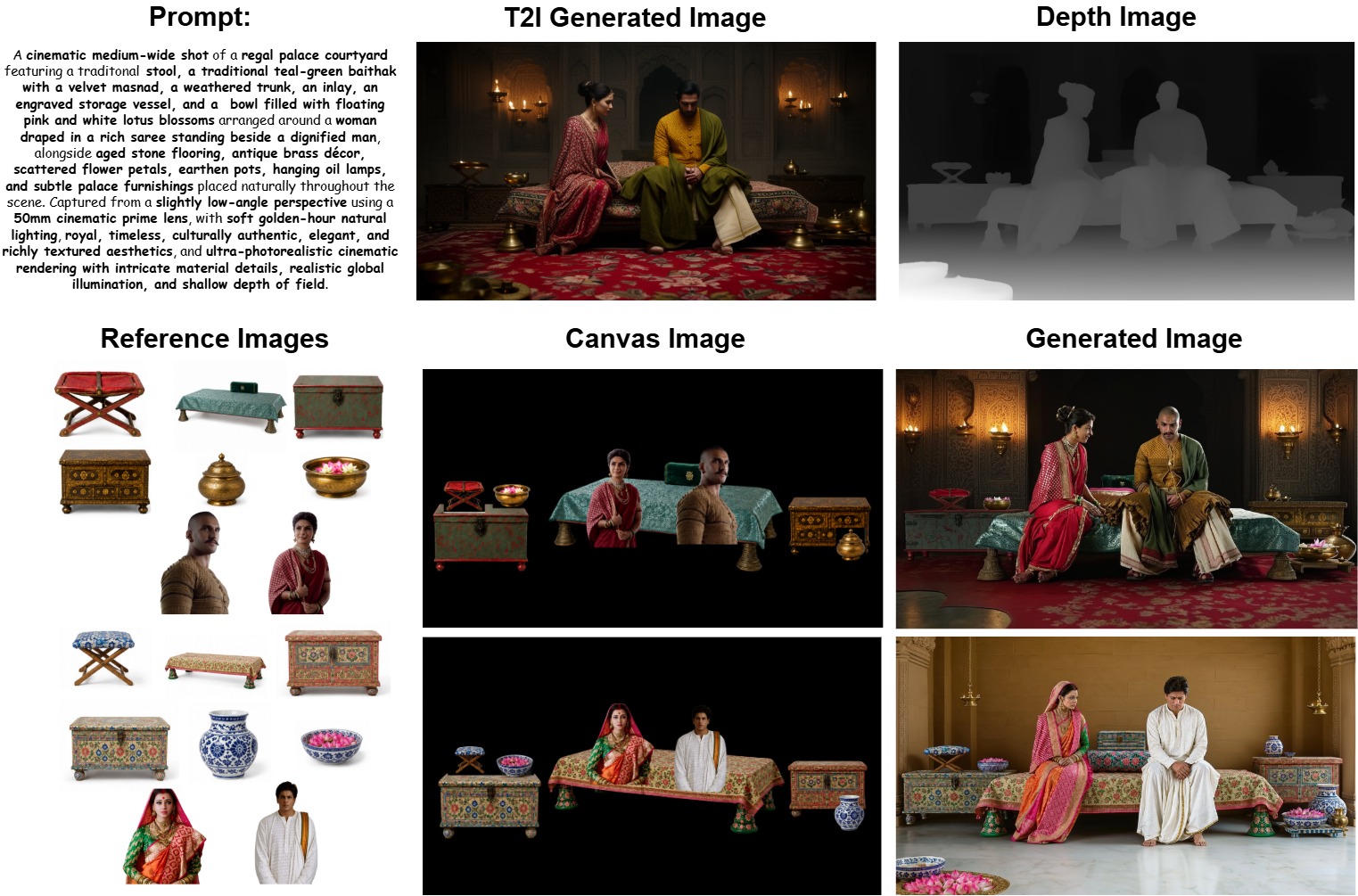}
  \captionof{figure}{Qualitative example with two different reference sets for the same scene layout. The prompt-induced canvas and aligned depth map provide shared structural priors, while the reference images specify subject appearance. RefCompose generates semantically consistent composites for both reference configurations.}
  \label{fig:multi_reference}
\end{center}

\paragraph{Depth-controlled orientation.}
Figure~\ref{fig:orientation_control} further isolates the role of the depth prior by holding the reference assets fixed and varying only the depth map.
In the first example, the depth map describes a conventional arrangement in which objects are naturally placed around a sweet shop.
In the second, the depth map is edited to depict a dynamic scene with floating debris, airborne bricks, and scattered objects.
Despite identical reference inputs, the generated results follow the geometry and orientation encoded by each depth map, demonstrating that RefCompose can control scene dynamics and spatial arrangement while preserving reference identity.
\begin{center}
  \includegraphics[width=\linewidth]{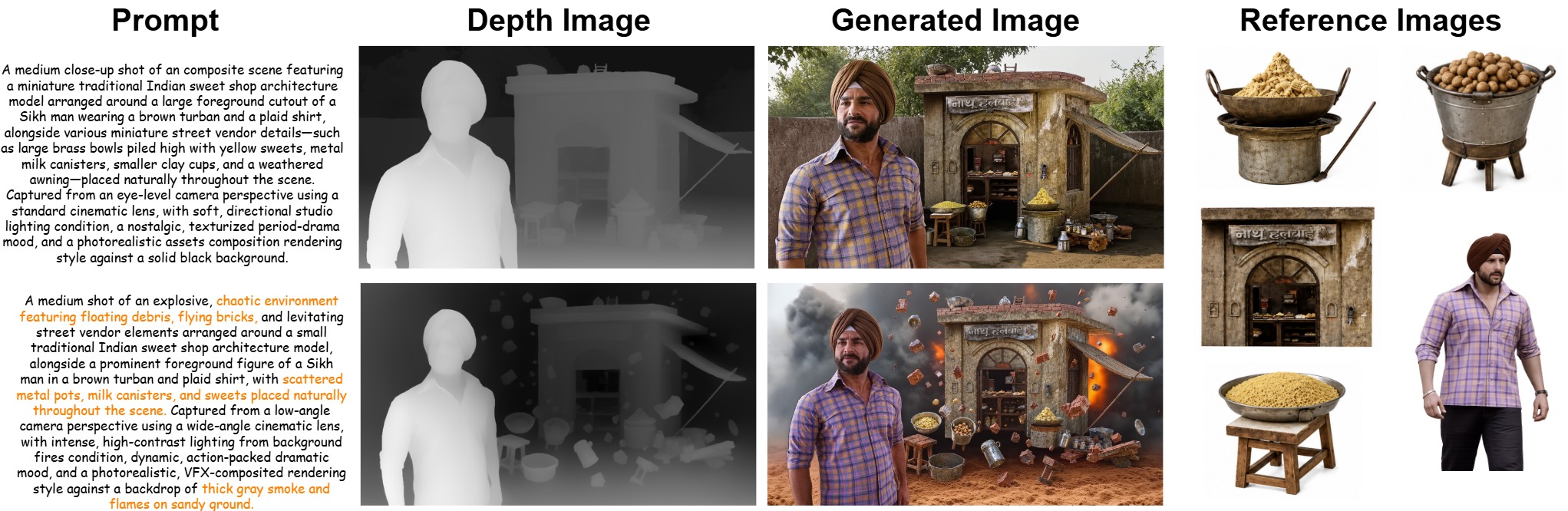}
  \captionof{figure}{Depth-controlled generation with identical reference assets. Changing only the depth map alters object orientation and scene dynamics, from a static sweet-shop arrangement to a chaotic environment with floating debris, while reference appearance remains stable.}
  \label{fig:orientation_control}
\end{center}

\subsection{Ablation Study}
\label{sec:ablation}
\begin{center}
  \includegraphics[width=\linewidth]{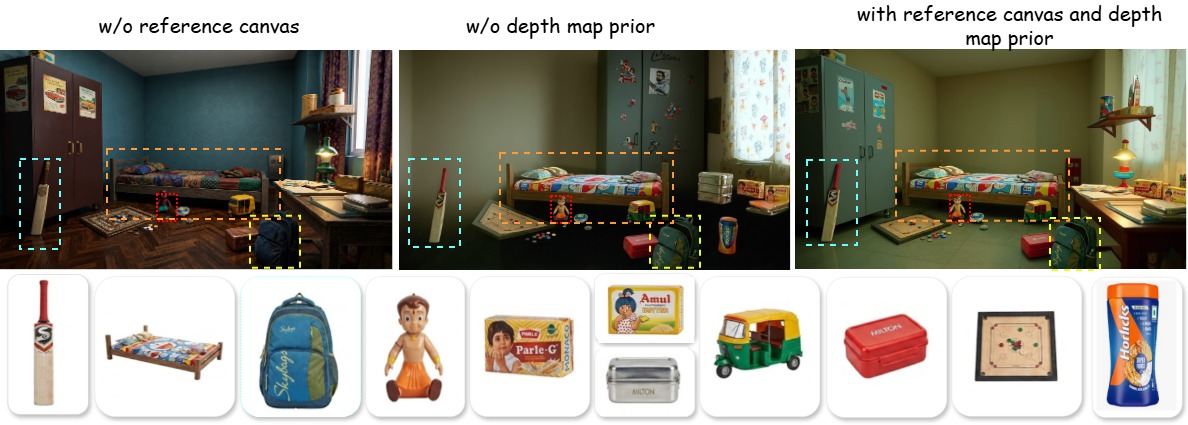}
  \captionof{figure}{Ablation study on an indoor multi-object scene. Removing either the reference canvas or depth prior degrades identity preservation or geometric consistency, while the full model preserves object appearance and spatial layout by leveraging both conditioning signals.}
  \label{fig:ablation}
\end{center}
Figure~\ref{fig:ablation} ablates the reference canvas and depth prior (reference crops shown above). Removing the canvas yields plausible layouts but hallucinated colours, textures, and identities; removing depth retains partial appearance yet degrades placement and occlusion (e.g., the bat and bed remain recognizable but are arranged implausibly). The full model preserves both appearance and geometry via decoupled LoRA streams; quantitative metrics are reported in the supplementary material.

\section{Conclusion}

This work addresses multi-reference image generation under a fixed memory cost, 
where token concatenation and QKV injection approaches struggle to preserve 
appearance fidelity and spatial precision, and text-only conditioning fails to 
maintain cultural and ethnic fidelity as the number of references and scene 
heterogeneity grow. RefCompose introduces a pixel-space compositional conditioning 
framework that encodes an arbitrary number of reference images onto a single 
fixed-resolution canvas, decoupling token complexity from reference cardinality. 
Spatial layout is induced from a structured text prompt through a frozen diffusion 
transformer, with Grounding DINO used to extract boxes from the generated layout 
rather than relying on annotated training boxes or a dedicated layout-generation 
model. A dual-stream LoRA injects depth and appearance signals through disentangled 
low-rank adaptation paths, enforcing geometric consistency and subject identity 
simultaneously. Quantitative evaluation demonstrates that RefCompose consistently 
outperforms reference-image baselines across appearance fidelity and spatial accuracy 
metrics, while maintaining constant inference memory regardless of reference count.

\section{Limitations}

RefCompose depends on a chain of upstream components, including prompt-induced layout $L$, Grounding DINO localisation, and a Depth Anything~3 map, whose errors compound when the structured template (Section~\ref{sec:layout-gen}) is ambiguous, when detections are missed or misaligned on generated images, or when monocular depth mis-orders overlapping regions on synthetic layouts.
Although canvas encoding avoids linear token growth with $N$, each reference is still resized to a fixed $H \times W$ patch, so large object counts or small assigned boxes attenuate fine-grained identity before encoding, and viewpoint mismatch between $L$ and the reference crops (e.g., bird's-eye depth with lateral portraits) can yield contradictory depth and appearance cues.
Our quantitative study caps previous methods at five references for fair comparison with memory-limited baselines, while reporting an additional RefCompose-only $N{>}5$ setting.

\bibliographystyle{splncs04}
\bibliography{main}
\end{document}